\documentclass[11pt,a4paper]{article}
\usepackage[hyperref]{ranlp2023}
\usepackage{times}
\usepackage{latexsym}

\hypersetup{
           breaklinks=true,   
           colorlinks=true,
        }

\usepackage{microtype}

\aclfinalcopy % Uncomment this line for the final submission
\title{cantnlp@LT-EDI-2023: Homophobia/Transphobia Detection in Social Media Comments using Spatio-Temporally Retrained Language Models}

\author{Sidney G.-J. Wong, Matthew Durward, Benjamin Adams \and Jonathan Dunn \\
  University of Canterbury, Christchurch, New Zealand \\
\texttt{\{sidney.wong,matthew.durward\}@pg.canterbury.ac.nz} \\
\texttt{\{benjamin.adams,jonathan.dunn\}@canterbury.ac.nz} \\ }
  
\date{}

\begin{document}
\maketitle
\begin{abstract}
This paper describes our multiclass classification system developed as part of the LT-EDI@RANLP-2023 shared task. We used a BERT-based language model to detect homophobic and transphobic content in social media comments across five language conditions: English, Spanish, Hindi, Malayalam, and Tamil. We retrained a transformer-based cross-language pretrained language model, XLM-RoBERTa, with spatially and temporally relevant social media language data. We also retrained a subset of models with simulated script-mixed social media language data with varied performance. We developed the best performing seven-label classification system for Malayalam based on weighted macro averaged F1 score (ranked first out of six) with variable performance for other language and class-label conditions. We found the inclusion of this spatio-temporal data improved the classification performance for all language and task conditions when compared with the baseline. The results suggests that transformer-based language classification systems are sensitive to register-specific and language-specific retraining.
\end{abstract}

\begin{table*}
\centering
\begin{tabular}{lcccc}
\hline
\textbf{Language Condition} & \textbf{H} & \textbf{N} & \textbf{T} & \textbf{Total}\\
\hline
English & 179 & 2978 & 7 & 3164\\
Hindi & 45 & 2423 & 92 & 2560\\
Malayalam & 476 & 2468 & 170 & 3114\\
Spanish & 200 & 450 & 200 & 850\\
Tamil & 453 & 2064 & 145 & 2662\\
\hline
\end{tabular}
\caption{\label{taska-training} The labelled training data broken down by language condition and class label for Task A. The class labels for Task A were homophobia (H), non-anti-LGBT+ content (N), and transphobia (T).}
\end{table*}

\begin{table*}
\centering
\begin{tabular}{lcccccccc}
\hline
\textbf{Language Condition} & \textbf{CS} & \textbf{HT} & \textbf{HD} & \textbf{HS} & \textbf{NO} & \textbf{TT} & \textbf{TD} & \textbf{Total}\\
\hline
English & 302 & 12 & 167 & 436 & 2240 & 1 & 6 & 3164\\
Malayalam & 152 & 57 & 419 & 69 & 2247 & 7 & 163 & 3114\\
Tamil & 212 & 37 & 416 & 218 & 1634 & 34 & 111 & 2662\\
\hline
\end{tabular}
\caption{\label{taskb-training} The labelled training data broken down by language condition and class label for Task B. The class labels for Task B were counter-speech (CS), homophobic-threatening (HT), homophobic-derogation (HD), hope-speech (HS), none-of-the-above (NO), transphobic-threatening (TT), and transphobic-derogation (TD).}
\end{table*}

\section{Introduction}
\label{sec:introduction}

The purpose of this shared task was to develop a classification system to predict whether samples of social media comments contained forms of homophobia or transphobia across different language conditions. There were no restrictions on language models or data pre-processing methods.

The five language conditions: English, Spanish, Hindi, Malayalam, and Tamil. In addition to the language conditions, participants were tasked with developing a system for a three-class and seven-class classification system defining different forms of homophobic and transphobic hate speech \cite{chakravarthi_dataset_2021}.

The main contribution of our proposed system outlined in this paper included spatio-temporal relevant social media language data to retrain a transformer-based language model to increase the sensitivity of the pretrained language model (PLM). We have also created simulated samples of script-mixed social media language data which was used as part of the retraining process.

\subsection{Problem Description}
\label{subsec:problem_description}

The organisers of this shared task provided .csv files containing labelled data of pre-processed comments of users reacting to LGBT+ videos on YouTube. This was an expanded data set of the Homophobia/Transphobia Detection data set \citep{chakravarthi_dataset_2021} with the inclusion of Hindi, Malayalam, and Spanish in addition to the pre-existing English and Tamil data. 

The comments were manually annotated based on a three-class and a seven-class classification system. The participants of the shared task were not provided any further information on the annotation process or measures of inter-annotator agreement. The shared task was broken down into the following tasks:

\begin{itemize}
\item Task A involves developing a classification model for three classes across all five language conditions as shown in Table \ref{taska-training}.
\item Task B involves developing a classification model for seven classes across three language conditions as shown in Table \ref{taskb-training}.
\end{itemize}

The organisers of this shared task provided training and validation data to develop the system. The test data was provided once the results of the shared task were announced. The organisers evaluated the performance of each homophobia/transphobia detection system with weighted macro averaged F1 score. The performance for each language and class-label condition were ranked based on this score.

\subsection{Related Work}
\label{subsec:related_work}

Previous approaches in detecting homophobia and transphobia on social media comments has shown varying levels of success \cite{chakravarthi_overview_2022}. In this shared task, participants were asked to detect homophobia and transphobia across three language conditions: English, Tamil, and an additional English-Tamil script-mixed data set.

Participants of the shared task combined various natural language processing methods such as statistical language models and machine learning to complete the task. However, the performance of transformer-based language models remained consistently high across all three language conditions. 

More specifically BERT-based models with minimal fine-tuning outperformed statistical language models using TF-IDF for feature extraction. BERT, or Bidirectional Encoder Representations from Transformers, structures the complex relationship between words in a language through embeddings \cite{devlin_bert_2019}.

The best performing BERT-based system for English yielded an average weighted macro F1 score of 0.92 compared with non-transformer-based language models \cite{maimaitituoheti_ablimet_2022}. Conversely, the same BERT-based models struggled to outperform machine learning and deep learning systems approaches in Tamil and in the English-Tamil condition.

This suggests further work is needed to refine BERT-based to improve its performance outside an English-context. Based on the promising results of BERT-based language models in  \citet{chakravarthi_overview_2022}, the current study extend on this transformer-based approach to develop and refine a  homophobia and transphobia detection system across language conditions.

\section{Methodology}
\label{sec:methodology}

In this section, we provide a system overview of our transformer-based language model. We also provide details on our retraining and fine-tuning procedures.

\subsection{System Overview}
\label{subsec:system-overview} 

Due to the number of language conditions for the current shared task, it was unfeasible to use language-specific BERT-based models. One risk for using independently developed language-specific BERT-based models was that there was no control on the source data used to train the representations. For this reason, we used a cross-lingual transformer-based language model as our baseline language model. 

XLM-RoBERTa was trained on two terabytes of CommonCrawl for 100 languages \cite{conneau_unsupervised_2020}. Some of these languages include English, Hindi, Malayalam, Spanish, and Tamil. Furthermore, Romanised Hindi and Tamil have also been included in the pretraining of this cross-lingual transformer-based language model.

Despite these benefits, we were aware of the risk in overgeneralising the register of language of CommonCrawl as the language used on this platform is not reflective of the language used on social media. We could retrain PLMs for a specific task to mitigate this issue without the need to train a PLM from scratch. 

This retraining method has shown to improve the performance of PLMs in downstream tasks (such as label classification) for under-represented and under-resourced languages by pretraining with additional register-specific language data \cite{liu_roberta_2019}. Therefore, we have retrained the baseline XLM-RoBERTa PLM prior to fine-tuning the baseline XLM-RoBERTa PLM.

\subsection{Retraining}
\label{subsec:retraining} 

\begin{table}
\centering
\begin{tabular}{lcc}
\hline \textbf{Language} & \textbf{Indic} & \textbf{Latin} \\ \hline
English & - & 50K \\
Spanish & - & 50K \\
Hindi & 50K & - \\
Malayalam & 50K & - \\
Tamil & 50K & - \\
SM Hindi & 37.5K & 12.5K \\
SM Malayalam & 37.5K & 12.5K \\
SM Tamil & 37.5K & 12.5K \\
\hline
\end{tabular}
\begin{tabular}{c}
\hline
\textbf{Total}\\
\hline
50K \\
50K \\
50K \\
50K \\
50K \\
50K \\
50K \\
50K \\
\hline
\end{tabular}
\caption{\label{corpus-size} Corpus size of language samples for fine-tuning with simulated script-mixing (SM).}
\end{table}

\begin{table*}
\centering
\begin{tabular}{lcccc}
\hline
\textbf{Language Condition} & \textbf{Baseline} & \textbf{Retrained} & \textbf{Script-Mixed}\\
\hline
English & 0.93 & \textbf{0.94} & - \\
Hindi & 0.93 & 0.92 & \textbf{0.97} \\
Malayalam & 0.93 & 0.95 & \textbf{0.94} \\
Spanish & 0.83 & (0.86) & - \\
Tamil & 0.70 & 0.93 & \textbf{0.93} \\
\hline
\end{tabular}
\begin{tabular}{c}
\hline
\textbf{Rank}\\
\hline
7 \\
3 \\
4 \\
- \\
3 \\
\hline
\end{tabular}
\caption{\label{taska-results} Macro averaged F1 for each language condition for Task A and overall rank for the shared task. The submitted result is in \textbf{bold}. Note that the result for Spanish was invalid.}
\end{table*}

\begin{table*}
\centering
\begin{tabular}{lcccc}
\hline
\textbf{Language Condition} & \textbf{Baseline} & \textbf{Retrained} & \textbf{Script-Mixed}\\
\hline
English & 0.15 & \textbf{0.54} & - \\
Malayalam & 0.86 & 0.86 & \textbf{0.88} \\
Tamil & 0.77 & 0.90 & \textbf{0.80} \\
\hline
\end{tabular}
\begin{tabular}{c}
\hline
\textbf{Rank}\\
\hline
6 \\
1 \\
4 \\
\hline
\end{tabular}
\caption{\label{taskb-results} Macro averaged F1 for each language condition for Task B and overall rank for the shared task. The submitted result is in \textbf{bold}.}
\end{table*}

We used social media language data from the Corpus of Global Language Use (CGLU) for retraining \cite{dunn_mapping_2020}. The CGLU is a very large digital corpora which contains over 20 billion words associated with 10,000 point locations across the globe.

Although the source of the CGLU social media language data comes from Twitter, a microblogging platform, and the training data comes from YouTube, a video sharing platform, our focus is on the written language components, and we assume some close domain alignment. We removed hashtags and hyperlinks to ensure the retraining data has a similar form to the training data. We also removed multiple punctuation and blank space characters. Short tweets with fewer than 50 characters were also systematically removed.

We controlled the spatial and temporal window of the sampled tweets by restricting the sample of tweets to those originating in India  produced between 1 January 2019 and 31 December 2019. Once again, we wanted to closely match retraining data with the time and geographic source of the labelled training data \cite{chakravarthi_dataset_2021}.

We used the {\small\verb|langdetect|}\footnote{https://pypi.org/project/langdetect/} library to detect the language condition for each tweet. For each of the five different language conditions, we extracted a random sample of 50,000 tweets for training. We then use the LanguageModelingModel class from the {\small\verb|simpletransformers|} library to retrain XLM-RoBERTa on an unlabelled corpus of social media language data.

In addition to creating corpus training data for the five different language conditions, we created additional corpus training data with simulated script-mixing. A major motivation to retrain the model with the simulated script-mixed retraining data is the lack of Romanised Malayalam in XLM-RoBERTa. We used the transliteration.XlitEngine class from the {\small\verb|ai4bharat|}\footnote{https://pypi.org/project/ai4bharat-transliteration/} library to transliterate one-fifth of the sample tweets from Indic to Latin script.

The size of our retraining corpora for each language condition is shown in Table \ref{corpus-size}. We retrained the language model for 4 iterations and we evaluated the training for every 500 steps. We saved the model with the best performance determined by the loss function in our output directory.

\subsection{Fine-tuning}
\label{subsec:finetuning}

Once we retrained XLM-RoBERTa with the social media language data from the CGLU, we fine-tuned the baseline and the retrained language models with the labelled training data.

As shown in Table \ref{taska-training} and Table \ref{taskb-training}, the class labels for both Task A and Task B are highly unbalanced. We used the RandomOverSampler class from the library to oversample the minority classes. In most cases, these minority classes related to homophobia and transphobia.

We used the classificaton class from the {\small\verb|simpletransformers|} library to fine-tune the retrained PLMs with the labelled training data. We trained the classification model for 8 iterations and we evaluated the training for every 500 steps. We also used AdamW optimization \cite{loshchilov_decoupled_2019}.

We applied the same fine-tuning strategy to Task A and Task B to maintain consistency across the shared task. We saved the model with the best performance determined by the loss function in our output directory.

\subsection{Other Settings}
\label{subsec:other_settings}

We completed the retraining and fine-tuning in Python3 on Google Colaboratory. We used GPU as our hardware accelerator using NVIDIA A100 Tensor Core graphics card.

\section{Results}
\label{sec:results}

The results of Task A are shown in Table \ref{taska-results} and the results of Task B are shown in Table \ref{taskb-results}. Both tables compare the weighted macro averaged F1 metrics for the classification models derived from the baseline XLM-RoBERTa and the modified XLM-RoBERTa models produced specifically for this task. The ranking of our models are also presented in the final column of the tables.

In Task A, English, Hindi, and Malayalam performed the best of the the baseline classification models with a macro averaged F1 score of .93. Tamil performed the worst of the baseline classification models. The performance of the retrained classification models were consistently better than the baseline classification models. Malayalam performed the best with a macro averaged F1 score of 0.95 while Spanish performed the worst with a macro averaged F1 score of 0.86. The classification models fine-tuned on simulated script-mixed training data did not improve the classification performance for Tamil. Conversely, we saw a decrease in performance for Malayalam. There was a large improvement in classification performance for Hindi.

We have highlighted the performance metric in bold in terms of the optimal classification models submitted to the organisers for evaluation in Table \ref{taska-results}. Hindi and Tamil ranked third out of seven, Malayalam ranked fourth out of seven, and English ranked seventh out of eleven. Due to issues with the labels, the submission for the Spanish condition in Task A was invalid. However, the macro averaged F1 metric is provided in brackets for reference.

In Task B, Malayalam performed the best of the baseline classification models with a macro averaged F1 score of 0.86 while English performed the worst with a macro averaged F1 score of 0.15. The performance increased once we fine-tuned the classification model with the retrained XLM-RoBERTa with the macro averaged F1 score for English improving from 0.15 to 0.54 and for Tamil improving from 0.77 to 0.90. The performance remained stable between the baseline and retrained models for Malayalam. 

When we introduced the script-mixed models for Malayalam and Tamil, we saw varying levels of performance. The macro averaged F1 score for Malayalam increased from 0.86 to 0.88. This suggests an increase in performance accuracy. Counterintuitively, the macro averaged score F1 for Tamil decreased from 0.90 to 0.80 which was on par with the baseline model. This suggests a decrease in performance accuracy.

Our Malayalam classification system fine-tuned on the script-mixed social media language data ranked first out of six, while the Tamil classification system fine-tuned on the script-mixed social media language data ranked fourth out of seven despite the decrease in performance from the retrained language model. Our English classification system fine-tuned on the retrained language model ranked sixth out of nine.

\section{Discussion}
\label{sec:discussion}

This paper addresses intrinsic issues related to hate speech detection in written social media data. This was mirrored in our training, validation, and testing data which reflects the multifaceted challenges of real-world scenarios. Hate speech is not confined to any single language or geographic region and its instances are often buried within the vast array of existing textual data, particularly in the context of social media. 

The employment of the XLM-RoBERTa model has demonstrated to be an effective system in detecting homophobia and transphobia in social media comments, particularly when the PLM has been retrained with spatio-temporal data for the English and Tamil language conditions. These findings underline the potential of integrating geographic language data into models as a means of enhancing their performance not only on highly represented languages, but also on lower underrepresented languages, thus offering a robust solution.

In the preceding sections of this paper, we have outlined and highlighted in Table \ref{taska-training} and Table \ref{taskb-training} an influencing challenge relating to the distribution of data across the different language conditions. The imbalance observed between language conditions as exhibited in the discrepancies in their respective training, validation, and test data sets poses an additional obstacle when it comes to drawing comparative inferences. However, there are opportunities that could help balance the data and potentially improve performance. 

To alleviate this issue, the utilisation of synthetic data through data augmentation techniques could prove to be a promising approach. Data augmentation, as a broad concept, involves expanding the existing data sets to enhance their diversity, and therefore, the generalisability of the models trained on them \cite{hoffmann_training_2022}. The generation of synthetic data has been demonstrated to be an effective mechanism in addressing biased data sets, but it also presents a desirable practice particularly suited for hate speech detection given the prevailing concerns over text obfuscation of such instances \cite{aggarwal_analyzing_2022}. 

To help facilitate a system that can account for these nuances, data noise injection via character, word, or even phrasal additions could be advantageous. In this sense, the application of synthetic data coupled with noise injection can help address class imbalance, while also training more robust classifiers that are less reliant on explicit instances of derogatory terms, but are more adept at discerning underlying contextual uses of hate speech. 

There are real-world applications to our homophobia/transphobia detection system as we can refine our model with language-specific and region-specific information to monitor hate speech on social media directed at LGBTQ+ communities. This is particularly useful for languages that are not otherwise as well represented in large language models such as Malayalam which saw great improvement in performance with the addition of script-mixed retraining data.

\section{Conclusion}
\label{sec:conclusion}

We saw an improvement in performance in our retrained homophobia/transphobia classification model when compared with our baseline model. Our unique approach to this shared task has shown potential for retraining pretrained language models with spatio-temporal relevant language data to improve the performance of our homophobia/transphobia detection system. Counterintuitively, the inclusion of script-mixed language data gave us variable results. We will aim to refine our classification system with other attested methods such as noise injection in order to improve the performance of our system.

\bibliographystyle{acl_natbib}
\bibliography{references.bib}

%\appendix

\end{document}